%% file: root.tex
\documentclass[letterpaper, 10 pt, conference]{ieeeconf}  

\IEEEoverridecommandlockouts                              

\usepackage{color}
\usepackage{caption}
\usepackage{graphics} 
\usepackage{amsmath} 
\usepackage{amssymb}  
\usepackage{biblatex} 
\usepackage{booktabs}
\usepackage{graphicx}
\usepackage[capitalise]{cleveref}

\DeclareMathOperator*{\argsortA}{arg\,sort}

\title{\LARGE \bf
Communication-Critical Planning \\via Multi-Agent Trajectory Exchange
}

\author{Nathaniel Moore Glaser and Zsolt Kira\\
Georgia Institute of Technology\\
\texttt{$\{$nglaser,zkira$\}$@gatech.edu}
}
\begin{document}

\maketitle
\thispagestyle{empty}
\pagestyle{empty}

\begin{abstract}
This paper addresses the task of joint multi-agent perception and planning, especially as it relates to the real-world challenge of collision-free navigation for connected self-driving vehicles.  For this task, several communication-enabled vehicles must navigate through a busy intersection while avoiding collisions with each other and with obstacles.  To this end, this paper proposes a \textit{learnable} costmap-based planning mechanism, given raw perceptual data, that is (1) \textit{distributed}, (2) \textit{uncertainty-aware}, and (3) \textit{bandwidth-efficient}.  Our method produces a costmap and uncertainty-aware entropy map to sort and fuse candidate trajectories as evaluated across multiple-agents.  
The proposed method demonstrates several favorable performance trends on a suite of open-source overhead datasets as well as within a novel communication-critical simulator.  It produces accurate semantic occupancy forecasts as an intermediate perception output, attaining a $72.5\%$ average pixel-wise classification accuracy.  By selecting the top trajectory, the multi-agent method scales well with the number of agents, reducing the hard collision rate by up to $57\%$ with eight agents compared to the single-agent version.
\end{abstract}

\input{introduction.tex}

\input{related_work.tex}
\input{method.tex}
\input{experiments.tex}
\input{conclusion.tex}

\printbibliography
\end{document}

%% file: introduction.tex
\section{Introduction}
For much of its history, robotics has revolved around single-agent systems.  
Many factors perpetuated this singular focus, such as costly hardware, weak compute power, and spotty inter-agent communication.  
However, thanks to modern technological advancements, these prohibitive developmental costs are largely reduced.  
With improved capabilities in hand, researchers have been keen to extend single-agent techniques towards their multi-agent counterparts.

One such extension involves that of deep learning~\cite{krizhevsky2012imagenet,simonyan2014very}.  This class of techniques can infer semantic information about the world by combining neural architectures with data.  It has seen remarkable success in areas such as object detection and, more recently, robotic perception and planning~\cite{DBLP:journals/corr/abs-2101-06547,zeng2019end,kollmitz2020learning,kollmitz2015time,chen2020interpretable}.  
However, again, these algorithms have primarily focused on \textit{single}-agent applications.  

To balance out this trend, this paper seeks to apply modern deep learning techniques towards the task of \textit{multi}-agent perception and planning, especially within the setting of collision-free trajectory planning for multiple connected robots.  
Specifically, we propose a \textit{distributed}, \textit{communication-enhanced} collision avoidance policy that is learned from observations of multi-agent interactions.  
We derive a novel trajectory exchange mechanism that leverages several state-of-the-art learning-based techniques, including image-to-costmap generation, deep-learned spatial uncertainty, and costmap-based motion planning.

To achieve this end, we distribute copies of an identical neural architecture across multiple connected agents.  Using this neural network, each agent then processes its local observations to determine (1) the cost of occupying different parts of its local state space (i.e. each agent generates a local costmap) and (2) the perceptual uncertainty associated with that cost (i.e. a entropy map).  We then propose a method whereby a querying agent broadcasts several motion trajectory candidates for distributed evaluation.  Then each supporting agent evaluates the incoming motion trajectories against its own egocentric costmap (and corresponding uncertainty map) and returns the sampled costs (and their uncertainties). Importantly, we propose an uncertainty-aware method to fuse the costs and uncertainties from multiple agents. Given the fused cost for each trajectory, the original agent then selects the lowest cost trajectory for execution. 

\input{figure/taskFigure.tex}
\input{figure/multiAgentArchitecture.tex}

We summarize the contributions of our paper as follows:
\begin{itemize}
    \item We address a variant on \textit{Multi-Robot Collaborative Perception and Planning}~\cite{liu2020when2com, liu2020who2com} in which a group of decentralized, but communication-enabled, agents must avoid collisions as they move through a shared world.
    \item To address this challenge, we propose an end-to-end learnable \textbf{M}ulti-\textbf{A}gent \textbf{T}rajectory \textbf{E}xchange network, \textbf{MATE}. We contribute novel techniques that extend the current state of the art, including a \textit{bandwidth-efficient}, \textit{uncertainty-aware} trajectory exchange mechanism.
    \item We evaluate our algorithm across a suite of open-source datasets~\cite{bock2019ind,autocast}, including within a novel, communication-critical simulator, \textbf{CoBEV-Sim}.
\end{itemize}

%% file: figure/taskFigure.tex
\begin{figure}
\vspace{1mm}
\centering
\includegraphics[width=\linewidth]{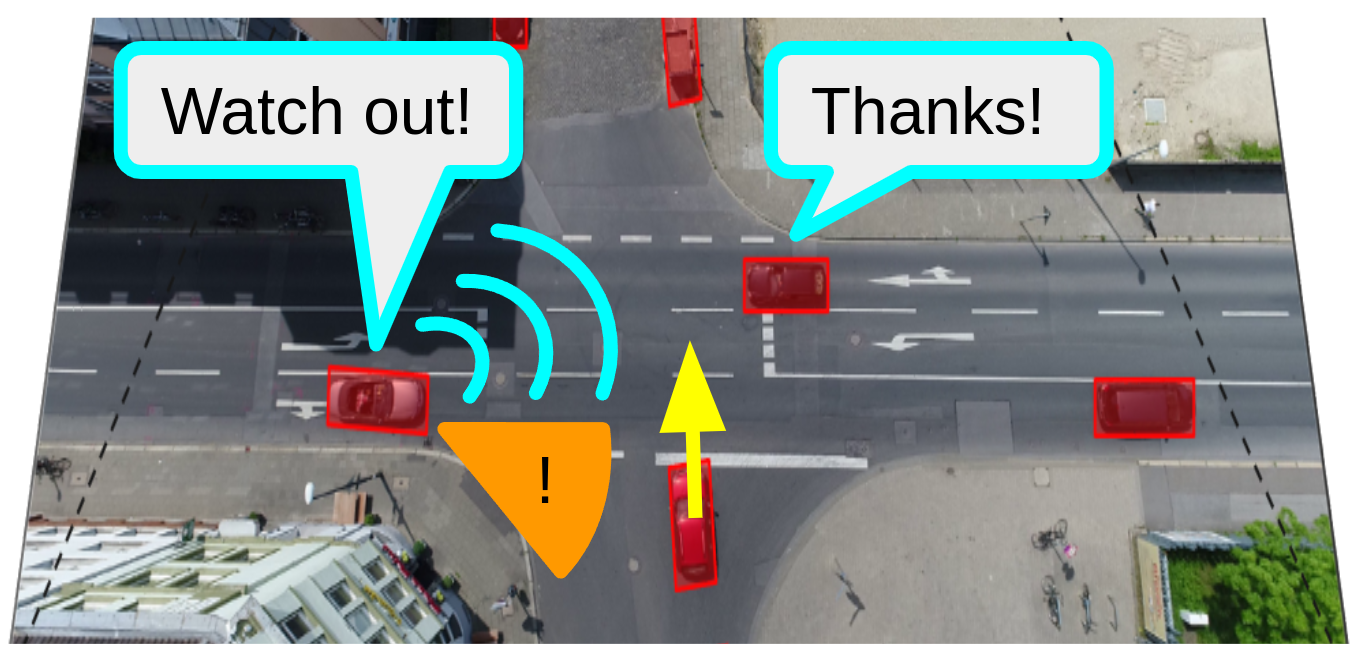}
\vspace*{-15pt}
\setlength{\belowcaptionskip}{-15pt}
\caption{\textbf{Task}. Collision-free planning via communication.}
\label{fig:TaskFigure}%
\end{figure} 

%% file: figure/multiAgentArchitecture.tex
\begin{figure*}
\vspace{2mm}
\centering
\includegraphics[width=0.8\linewidth]{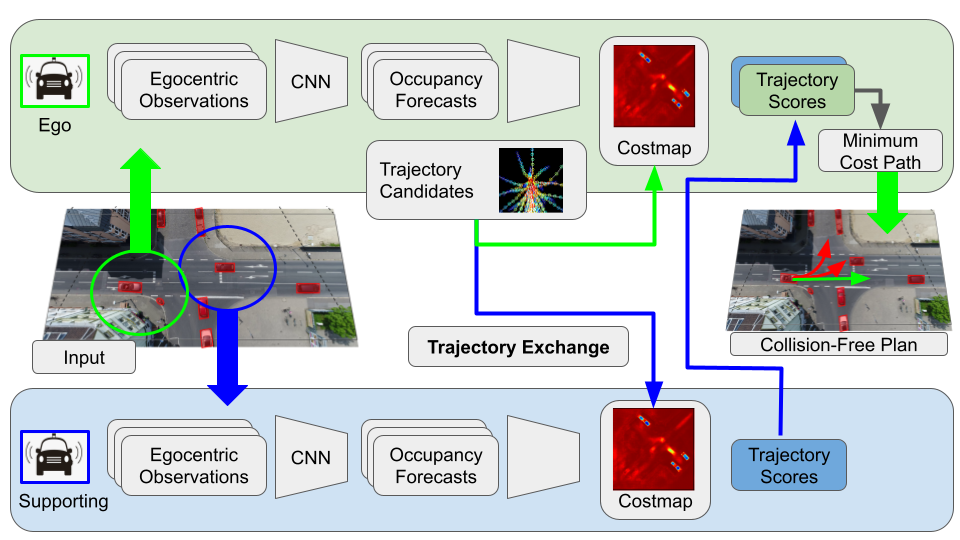}
\setlength{\belowcaptionskip}{-15pt}
\caption{\textbf{Multi-Agent Trajectory Exchange}.  Egocentric observations from multiple connected vehicles are locally processed into learnable and interpretable costmaps.  Each agent then broadcasts its trajectory candidates for distributed, opportunistic evaluation across any number of supporting agents.  The supporting agents then return the costs sampled by the trajectory.  The ego agent aggregates these scores using uncertainty-aware fusion and selects the minimum cost trajectory for execution.}
\label{fig:multiAgentArchitecture}%
\end{figure*}

%% file: related_work.tex
\section{Related Work}
Several prior works are particularly relevant for collaborative perception and planning.

One set of prior work~\cite{kiran2021deep,sadat2020perceive,kollmitz2015time,kollmitz2020learning} addresses how a motion planning policy may be learned from data, either by interacting with an environment (reinforcement learning) or by observing an expert's interactions with an environment (inverse reinforcement learning).  
The work of Sadat et al.~\cite{sadat2020perceive} shows how a neural motion policy may be learned by observing expert driving demonstrations.  Their system takes a map and lidar sequence as inputs, and it uses a learned encoder to produce a set of occupancy predictions (e.g. pedestrian, car, bike) for future timesteps.  Next, it uses a learned cost function to score a set of sampled trajectories against these occupancy predictions, and the minimum cost trajectory is selected for execution.  

Other, more human-centric, works consider the interplay between multiple human agents in a shared scene.  For instance, the works of Kollmitz et al.~\cite{kollmitz2015time, kollmitz2020learning} leverage physical interaction and inverse reinforcement learning to adapt the planning behavior of a robot near human actors.  However, while the robots in these works explicitly consider the motion of other actors, they do not leverage communication to improve their capabilities.

To this end, another set of prior work has addressed the challenge of \textit{collaborative perception} through learned communication~\cite{liu2020who2com, liu2020when2com, wang2020v2vnet}.  Learned communication is a relatively recent trend where data from different sensing sources is pegged for exchange, based on a purely learned process.  \textbf{When2Com}~\cite{liu2020when2com} leverages a learned \textit{handshake communication} module to mediate the exchange of visual information between robotic agents, specifically for multi-view image classification and semantic segmentation.  

A similar set of prior work has addressed collaborative perception through learned \textit{transformation} and \textit{communication} of spatial feature maps~\cite{Li_2021_NeurIPS,glasermash,DBLP:journals/corr/abs-2101-06547}.  For instance, \textbf{Disconet}~\cite{Li_2021_NeurIPS} uses a convolutional network to compress top-down 2D lidar rasters into an intermediate feature representation.  The spatial cells inside these feature maps are then synchronized across multiple agents using a spatial collaboration graph.  

Using similar communication architectures, more recent works have addressed the joint task of \textit{collaborative perception and planning}.  \textbf{Coopernaut}~\cite{coopernaut} uses a learned encoder module to compress point cloud information into a relatively compact 2D representation.  The 2D representations from multiple allocentric agents are then transmitted, spatially transformed, and aggregated into the perspective of a single egocentric agent.  Finally, a learned control module executes a control policy based on the aggregated information.  However, these works only implicitly consider the role of uncertainty when aggregating multi-agent information.

An exciting new trend involves leveraging deep-learned uncertainty~\cite{kendall2017uncertainties,pan2020domainadaptation,wang2017activelearning,tian2020uno}.  Pan et al.~\cite{pan2020domainadaptation} argue that the softmax confidences produced by classification networks roughly correlate with the empirical accuracy, as evaluated across a diverse set of common classification tasks.  Converse to confidence, several works~\cite{wang2017activelearning, pan2020domainadaptation,meng2021spatialuncertainty} apply Shannon Entropy~\cite{shannon2001} as a measure of \textit{uncertainty}, and demonstrate its application to tasks such as domain adaptation, active learning, and crowd counting.  Moreover, other works~\cite{meng2021spatialuncertainty,uhl2018spatialisinguncertainty,tian2020uno} leverage entropy as a spatial map, especially for semantic segmentation tasks.  Wang et al.~\cite{wang2018entropyweightedfusion} uses an entropy-weighted average for image fusion.

Our approach draws inspiration from each of these bodies of work.  Like prior work, our method uses an interpretable costmap-based scoring mechanism for evaluating and selecting trajectories for execution.  However, unlike prior work, our method uses a highly efficient \textit{trajectory} exchange mechanism to improve planning performance, instead of transmitting dense representations (such as spatial feature maps).  Moreover, our exchange mechanism explicitly considers perceptual uncertainty as an intuitive way to fuse the trajectory scores from multiple robot sources.  

Beyond extending the methodology of prior work, our work also advances the setting of collaborative perception and planning.  Several prior works add their own specialized datasets~\cite{Li_2021_RAL,autocast} to the corpus of existing open-source datasets~\cite{dosovitskiy2017carla, suo2021trafficsim, howe2021weakly, yu2022dair} and closed-source datasets~\cite{wang2020v2vnet, chen2019cooper, manivasagam2020lidarsim, xiao2018multimedia, maalej2017vanets}.  \textbf{V2X-Sim}~\cite{Li_2021_RAL} primarily focuses on the challenge of collaborative perception, though it only offers relatively simple planning scenarios.  For our work, we evaluate our algorithm on a dataset split of \textbf{AutoCastSim}~\cite{autocast}.  This dataset specifically targets challenging self-driving situations, such as red light infringements and left turns across traffic.  However, these scenarios lack diversity and have a limited number of agents available for useful communication.  To address these limitations, our work extends the capabilities of these datasets with the novel \textbf{CoBEV-Sim}, a simulator which creates a higher volume of diverse, communication-critical scenarios.

%% file: method.tex
\section{Methodology}
Autonomous navigation is an important capability for self-driving vehicles (SDVs).  Prior to any widespread adoption, SDVs must first demonstrate a robust ability to move safely in a cluttered, dynamic environment.  To this end, we propose a novel approach for performing \textit{safe} autonomous navigation, especially in the context of inter-vehicle communication.

\subsection{Task}
This work addresses the task of collision-free intersection navigation for connected robotic agents.  Namely, given pose in a global coordinate frame, local perception, and inter-agent communication, a group of multiple autonomous robots must plan a sequence of actions that limits dangerous interactions with the environment and with each other, as shown in Fig.~\ref{fig:TaskFigure}.  

In this setting, each agent has access to three types of data: (1) global pose data, (2) local perception data, and (3) data communicated from nearby agents.  For the global pose data, each vehicle has knowledge of its past (and current) positions and orientations with respect to a global coordinate system (as provided by GPS or similar).  For the local perception data, each vehicle captures \textit{egocentric} visual information as it travels through the environment.  For the communicated data, each vehicle may communicate bandwidth-limited information (as measured in bytes) to other vehicles within a limited range.  These limitations are typically dictated by real-world environmental constraints.  This constrained communication provides the agents with an indirect ability to \textit{perceive} beyond their local observations and \textit{plan} for future interactions.  This setting is further complicated by the presence of both communicating and non-communicating mobile agents.

\subsection{Model Architecture}
As summarized in Fig.~\ref{fig:multiAgentArchitecture} and~\ref{fig:qualitativeSummary}, this work leverages several modules to tackle the task of intersection navigation:

\textbf{Motion Forecasting}. First, as with Chen et al.~\cite{chen2020interpretable}, each agent uses a CNN $\Phi$ to convert a sequence of aligned 2D Bird's Eye View (BEV) observations $\textbf{X}^{T^-:0}$ into a sequence of 2D semantic occupancy forecasts $\textbf{P}^{1:T^+}$:   
\begin{equation}\label{eq:modelprediction}
  \textbf{P}_{\{\text{null},\text{ego},\text{allo}\}}^{1:T^+} = \Phi(\textbf{X}^{T^-:0})
\end{equation}
The output classes include a \textit{null}, egocentric (\textit{ego}), and allocentric (\textit{allo}) agent class, trained via cross-entropy loss.

\textbf{Costmap and Uncertainty-Map Generation}. Next, each agent converts its 2D semantic occupancy forecasts into a costmap and entropy map.  The costmap is created by combining its egocentric $\textbf{P}_{ego}$ and allocentric $\textbf{P}_{allo}$ occupancy forecasts into a binarized occupancy map \textbf{B} using threshold $\theta$ and then computing a signed distance field (\textbf{SDF})~\cite{oleynikova2016signed}:
\begin{equation}\label{eq:sdf}
  \textbf{B}^t = \max{(\textbf{P}_{ego}^t, \textbf{P}_{allo}^t)} > \theta
  \quad\text{and}\quad
  \textbf{D}^t = \textbf{\text{SDF}}(\textbf{B}^t)
\end{equation}
To generate the uncertainty map, Shannon entropy is computed across the softmaxed class channels for each spatial cell in the semantic occupancy forecast\footnote{In this section, we denote image-structured data with bold uppercase (e.g. $\textbf{U}^t$) and extracted cells in lowercase (e.g. $u_{x,y}^t$).}, similar to the procedures described in~\cite{meng2021spatialuncertainty,uhl2018spatialisinguncertainty,tian2020uno}:
\begin{equation}\label{eq:entropy}
  u_{x,y}^{t} = -\sum_{c}{p_{x,y}^{t}(c) \log{p_{x,y}^{t}(c)}}
\end{equation}

\textbf{Trajectory Generation}. Each agent generates a set of $N$ trajectory candidates, $\mathbb{S} = \{S^{i}\}_0^N$, where each trajectory candidate defines a sequence of agent poses at future timesteps,  $S^{i} = \{(x_t,y_t)\}_1^{T^+}$.  The trajectories were generated to cover a wide range of feasible vehicle maneuvers, as dictated by linear and angular acceleration limits.

\textbf{\textit{Bandwidth-Efficient} Trajectory Exchange}. Each ego agent ($e$) broadcasts its candidate trajectories for distributed evaluation by the supporting agents ($a$).  Each supporting agent transforms these trajectories to its local frame using relative pose matrix $R_e^a$.  Next, each supporting agent returns values sampled from its costmap and entropy map:
\begin{equation}\label{eq:trajectorysampling}
  S_t^{a,i} = R_e^a S_t^{e,i} = (x(t), y(t))
\end{equation}
\begin{equation}\label{eq:trajectorysampling}
    C_{t}^{a,i} = \textbf{D}_{t}^{a,i}[x(t),y(t)]
\end{equation}
\begin{equation}\label{eq:trajectorysampling}
    U_{t}^{a,i} = \textbf{U}_{t}^{a,i}[x(t),y(t)]
\end{equation}

\textbf{\textit{Uncertainty-Aware} Multi-Agent Cost Fusion}. After having its suite of candidate trajectories scored by the cooperating agents, the ego agent fuses all of these contributions together via an entropy-weighted average, similar to~\cite{wang2018entropyweightedfusion}:
\begin{equation}\label{eq:multiagentfusion}
  F^{i} = \sum_{a \in \text{allo}} \sum_{t = 1}^{T^+} \frac{C_{t}^{a,i}}{U_{t}^{a,i}}
\end{equation}

\textbf{Trajectory Prioritization}.  Finally, the set of trajectories is sorted based on their fused scores:
\begin{equation}\label{eq:multiagentfusion}
    \mathbb{S}_{\text{prioritized}} = [S_i | \forall i \in \argsortA_i F^i]
\end{equation}

\input{figure/qualitativeSummary.tex}

\subsection{Dataset}
Our problem domain combines several distinct research directions (motion forecasting, motion planning, communication-criticality, and safety for self-driving vehicles) all within a collaborative multi-robot setting.  Datasets exist for each of these domains; however, it is challenging to find a single dataset that fairly unifies all of them.  Therefore, we evaluate our approach using both slightly modified open-access datasets as well as a novel communication-critical simulator.  We seek to demonstrate that, in certain rare (but important!) scenarios, having access to a team of collaborating agents (and their local perceptions) improves planning performance.

Namely, we include the Intersection Drone Dataset (\textbf{InD}), \textbf{AutoCastSim}, and our own Collaborative Bird's Eye View simulator (\textbf{CoBEV-Sim}) in our analysis.  Each of these datasets is adapted to the same general task: a set of communication-enabled agents must safely maneuver through a shared environment that also includes a set of non-communicative agents.

Specifically, we adapt each dataset to a uniform style of egocentric \textbf{BEV} inputs and outputs\footnote{The egocentric BEV perspective is the output of prior research works~\cite{Li_2021_RAL}.  For simplicity, we evaluate our algorithm using this data format as a starting point.}.  For each communication-enabled agent, we capture a limited-range, top-down perspective of the shared environment, centered and oriented with respect to that agent.  These top-down perspectives can take the form of a 2d lidar scan, lidar scan raster, and local segmentation mask (i.e. occupancy mask), and we capture them at a constant synchronized frame rate throughout the duration of each planning scenario.  Moreover, we record the relative pose between each pair of agents, communicating and non-communicating (the poses of the non-communicating agents are solely used for evaluation).  We separate the observed sequence of top-down rasters and relative poses into two categories -- past data and future data.  The past information is used as input, whereas the future information is used for training and evaluation.

The ultimate task of each dataset is the following: each communicating agent observes a sequence of past, egocentric occupancy masks; then each agent must efficiently leverage the inferences of its peers to propose a set of trajectory candidates that are collision-free for some future time horizon.

\subsubsection{Open-Access Datasets}

We port several open-access datasets to our unified collaborative perception and planning setting.  Each of these datasets includes motion trajectories for scenarios with many cooperating (i.e. non-colliding) agents.  The \textbf{InD} dataset consist of extensive real motion tracks of interacting road users from an overhead \textbf{BEV} drone perspective.  The \textbf{AutoCastSim} dataset consists of the same data, though synthesized within the CARLA simulator.

\subsubsection{Communication-Critical Simulator}

In addition to open-access datasets, we evaluate the performance of our approach within a novel, communication-critical simulation environment.  This simulator addresses several shortcomings observed across many available datasets:

\textbf{Existing datasets are cumbersome.}  They often require a bulky simulator, complex API calls, or heavy downloads.

\textbf{Existing datasets lack diversity.} They support relatively few distinct scenarios, offering insignificant scene diversity.

\textbf{Existing datasets cover inconsequential settings.} There is an imbalance between common, benign situations (like lane-following) and rare, dangerous ones (like blind corners).

\textbf{Existing datasets do not contain interoperable agents.} Most datasets focus on a single "ego" agent, instead of allowing any agent to become a benefactor of communication.

Our dataset addresses each of these concerns.  It is \textbf{lightweight}, \textbf{randomizeable}, and \textbf{customizeable}.  It is straightforward to seed situations which are interesting, especially situations which intuitively would benefit from multi-agent communication.  Moreover, each agent can be ``put into the driver's'' seat, and its planning performance can be fairly and uniformly evaluated.

Specifically, our simulator allows for the dynamic generation of communication-critical multi-agent planning scenarios.  First, a user specifies the start and goal locations of any number of agents.  These agents can be spawned with a customizeable footprint, either corresponding to mobile road users (such as a car, truck, or pedestrian) or other communication-enabled road infrastructure (such as a smart traffic light or building).  Next, the simulator iteratively assigns a randomized straight-line speed profile to each agent such that it avoids collisions with its neighbors.  The assignment process is repeated until all agents have a non-colliding, goal-reaching trajectory sequence.  Finally, the simulator then renders the local perception data (2D lidar scan, BEV lidar raster, and BEV semantic segmentation map) for each of these agents.  These lidar scans take into account realistic limitations, such as sensor range limitations and perceptual occlusions.

Using our simulator, we generate a set of communication-critical scenarios: 

\textbf{CrissCross}. This scenario consists of many straight-line trajectories that chaotically intersect in open space.  For each scenario, we randomize the start and goal positions of $20$ agents.  A random subset of these agents is communication-enabled ({$n_{com} = \{1, 5, 10, 15\}$} for our experiments), while the remaining agents are non-communicative ($n_{silent} = 20 - n_{com}$).  This scenario is communication-critical because agents frequently transit through the paths of each other, oftentimes from outside of their egocentric perception range; the communicating agents must rely on collaborators to enhance their effective observation range.

\textbf{AutoCastSim}. This scenario includes a mixture of the red light infraction and blind corner cases from the original open-access dataset.  Since the original dataset includes only a small number of nearby collaborating agents ($\sim3$), we randomly insert an additional set of static and trailing agents to the planning scene.  We include $8$ total agents, with a mix of communicating agents ({$n_{com} = \{1, 2, 4, 8\}$}) and non-communicating agents ($n_{silent} = 8 - n_{com}$).  These supplementary agents undergo the same collision-free randomization described earlier, ensuring novel scenarios after each randomization.

We leverage the randomization component of our simulator during training and evaluation.   During each training epoch, we synthesize $10$ novel planning scenarios.  For evaluation, we use randomization to create a static test suite of $100$ planning scenarios.

%% file: figure/qualitativeSummary.tex
\begin{figure}
\centering
\includegraphics[width=\linewidth]{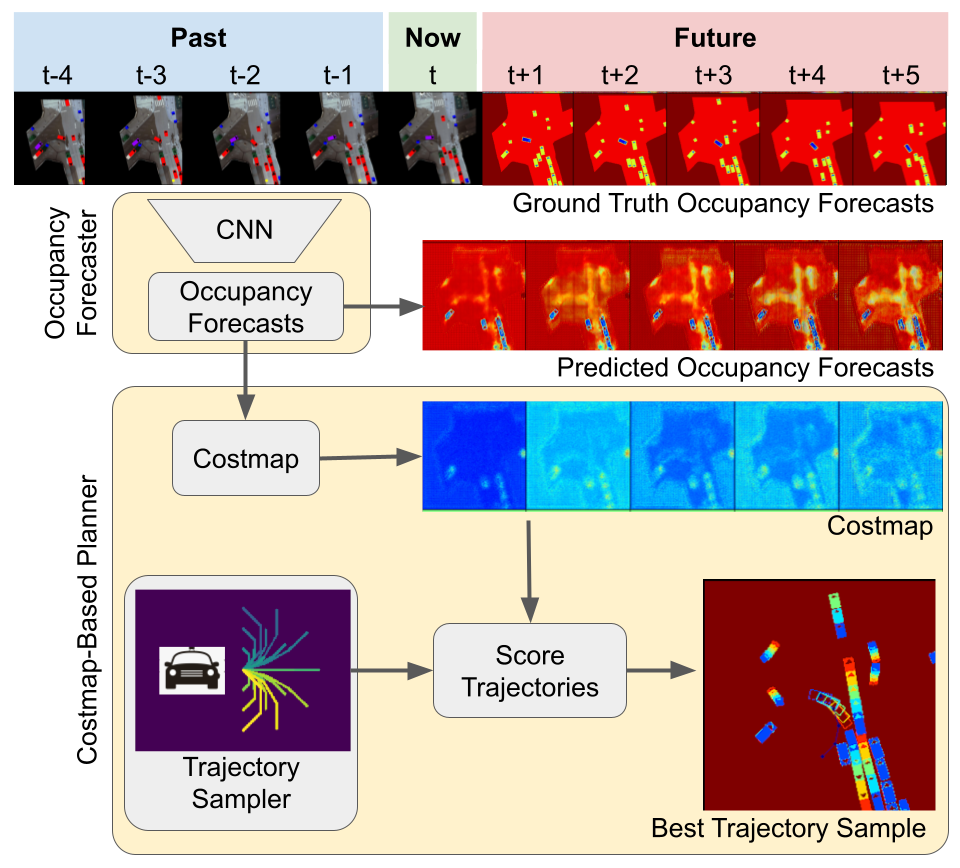}
\vspace*{-15pt}
\setlength{\belowcaptionskip}{-15pt}
\caption{\textbf{Overview of qualitative results for a single agent}.  
}
\label{fig:qualitativeSummary}%
\end{figure}

%% file: experiments.tex
\subsection{Experiments}
We evaluate the performance of our proposed model on the \textbf{InD}, \textbf{AutoCastSim}, and \textbf{CoBEV-Sim} datasets.  Our model uses an input BEV sequence to infer future segmentation masks and to produce a candidate trajectory output.  To train the network, these inferred segmentation masks are compared against the ground truth segmentation masks using a cross entropy loss function.  To test the network, the inferred segmentation masks are compared against the ground truth masks to yield accuracy and mean intersection of union (mIoU) metrics.  To test planning performance, the trajectories are ``rolled-out'' in the dataset environment, where potential collisions with the environment and its obstacles are tallied.  Both qualitative and quantitative assessments are discussed in the following section.  

\section{Results}
The results presented in the beginning of this section highlight the performance of single-agent perception and planning, as demonstrated on the \textbf{InD} dataset.  Multi-agent perception and planning is discussed towards the end of the section.  Those results are evaluated on the \textbf{InD}, \textbf{AutoCastSim}, and \textbf{CoBEV-Sim} datasets.
 
\input{figure/motionForecastResults.tex}
\input{figure/quantitativeResults.tex}

\subsection{Motion Forecasting}
We assess perception performance by comparing the inferred occupancy forecasts against ground truth future occupancy.  Qualitatively, as shown in Fig~\ref{fig:motionForecastResults}, the predicted locations of various road users roughly matches what actually unfolded in the scene.  In other words, the occupancy heat maps for the egocentric agent and dynamic objects constitute reasonable hypotheses for where different road actors may be found at different points in the future.  Moreover, the model seems to have learned that moving objects maintain their speed and heading, within some variability.  For instance, the motion forecasting module predicted a diagonal streak of probable locations for the egocentric agent---since the agent was previously travelling in a straight line at constant speed, it will likely continue on that path.  The streak is likely stretched in the longitudinal direction to account for uncertainty in velocity.  There is less stretching in the transverse direction likely because the road surface constrains that variability.  Additionally, the model is able to learn turning behavior and multi-modal behavior, as shown in Fig~\ref{fig:qualitativeSummary}.

Quantitatively, the predicted and ground truth segmentation forecasts are compared using standard accuracy and mean intersection over union (mIOU) metrics.  As shown in Table~\ref{tab:quantitativeResults}, the motion forecasting module achieves an average pixel-wise accuracy of $72.5\%$ and mean IoU of $50.9\%$.  As expected, the performance of the motion forecasting module degrades for temporally distant forecasts.

\subsection{Single-Agent Trajectory Sorting}
The proposed method infers a cost that allows for a set of trajectories to be scored and prioritized.  An indirect way to measure the quality of the cost function (and hence the overall learned pipeline) is to see how sorting trajectories by their cost correlates with key metrics, such collision rate.  As shown in Fig~\ref{fig:rankingCosts}, the inferred trajectory cost correlates with collision rate.  

\input{figure/rankingCosts.tex}
\input{figure/collisionRateTable.tex}

\subsection{Multi-Agent Trajectory Sorting}

We evaluate our method on the communication-critical datasets of \textbf{AutoCastSim} and \textbf{CoBEV-Sim}.  These datasets include important scenarios where communication is crucial to improving safety, such as maneuvering past a blind corner at high speed.  We present our results in terms of top-$k$ collision rates.  The top-$k$ collision rate denotes the frequency of a hard collision as averaged across the $k$ trajectories that have the lowest cost, as inferred by our multi-agent trajectory exchange method.  Oftentimes, several trajectory candidates can safely maneuver through a scenario.  We include these top-$k$ metrics to capture the performance of both the best trajectory ($k=1$) and the best \textit{set} of trajectories ($k=10$).

As shown in Table~\ref{tab:collisionRateTable}, our method highlights several desirable trends.  First, the top-$1$ trajectory outperforms the top-$10$ trajectories, and both outperform the random trajectory sampler baseline.  This trend indicates that our learned, distributed trajectory prioritization scheme is properly inferring the cost of trajectory samples, proportional to empirical collision rates.  Second, the collision rate improves with more inter-agent communication.  The top-$1$ trajectory shows a $57.0\%$ decrease and a $29.7\%$ decrease from the single agent collision rate for \textbf{AutoCastSim} and \textbf{CoBEV-Sim}, respectively.  Third, the top-$1$ and top-$10$ collision rates are relatively similar, indicating that our method has prioritized trajectories such that downstream tasks can safely choose between the top-$k$ set of trajectories with minimal penalty, a desirable feature for more downstream planners.  Finally, communication is critical in the \textbf{CoBEV-Sim} dataset, as shown by the similar collision rates in the non-communicating case ($n=1$), where single-agent performance matches that of a random trajectory sampler.

For the \textbf{InD Dataset}, our multi-agent trajectory exchange mechanism (with support from about $4$ agents) yielded a $7.2\%$ decrease in hard collisions over its single-agent counterpart.  The solid and dotted lines of Fig~\ref{fig:rankingCosts} show the close relative performance of the multi-agent and single-agent algorithms.  This improvement is likely a result of communication being non-essential for this specific dataset: a single independent agent can easily handle most \textbf{InD} scenarios. 

\subsection{Bandwidth}

Several prior works~\cite{Li_2021_NeurIPS,coopernaut} address multi-agent collaborative perception by exchanging dense representations of perception information.  These representations, while compressed, still consume significant amounts of bandwidth, especially since they preserve 2D image structure.  Our work takes bandwidth-efficiency to the extreme.  For instance, if all agents share a common dictionary of trajectory samples (i.e. $80$ trajectory samples with $15$ 2D pose waypoints), then the egocentric agent only needs to transmit its relative location (i.e. a single $(x,y,\theta)$ pose); and the allocentric agents can roto-transform their shared dictionaries of trajectories to the egocentric perspective and simply broadcast the values that were sampled from its costmap and uncertainty map.  This 1D list of sampled values ($80 \times 15 \times 2 = 2400$ floating point values) is significantly more compressed than the 2D image-based representations of prior work (i.e. DiscoNet~\cite{Li_2021_NeurIPS} with $\sim10^{4}$ floating point values).  

%% file: figure/motionForecastResults.tex
\begin{figure}
\vspace{2mm}
\centering
\includegraphics[width=\linewidth]{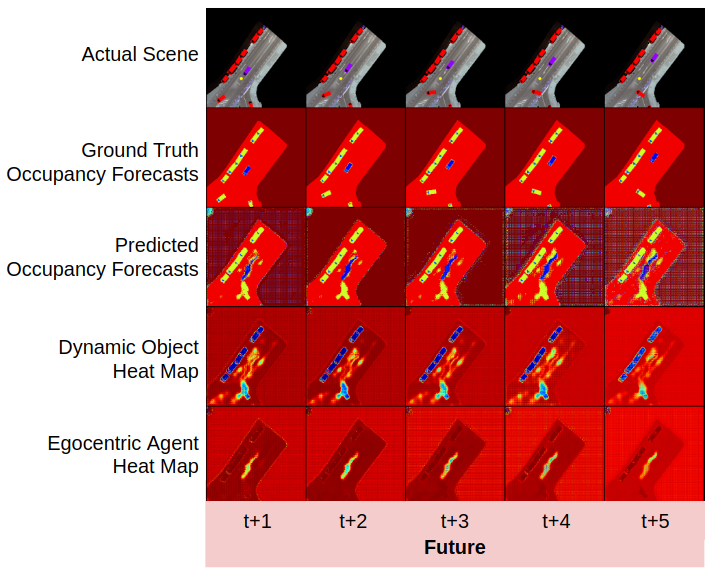}
\vspace*{-15pt}
\setlength{\belowcaptionskip}{-15pt}
\caption{\textbf{Sample qualitative results for motion forecasting}.  Using a past sequence of egocentric data (not pictured), the motion forecasting module predicts future state occupancy. 
}
\label{fig:motionForecastResults}%
\end{figure} 

%% file: figure/quantitativeResults.tex
\begin{table}
\vspace{5mm}
\resizebox{\linewidth}{!}{%
\begin{tabular}{l|ccccc|c}
&\multicolumn{5}{c}{ForecastTime} & \\
& $t=1$ & $t=2$ & $t=3$ & $t=4$ & $t=5$ & $avg$ \\
\hline
Mean Accuracy & 78.0 & 78.3 & 73.0 & 68.2 & 65.0 & 72.5 \\
Mean IoU & 49.3 & 53.4 & 53.1 & 51.1 & 47.5 & 50.9 \\
\end{tabular}
}
\setlength{\belowcaptionskip}{-15pt}
\caption{\textbf{Segmentation metrics for motion forecasting.}
}
\label{tab:quantitativeResults}
\end{table}

%% file: figure/rankingCosts.tex
\begin{figure}
\vspace{2mm}
\centering
\includegraphics[width=\linewidth]{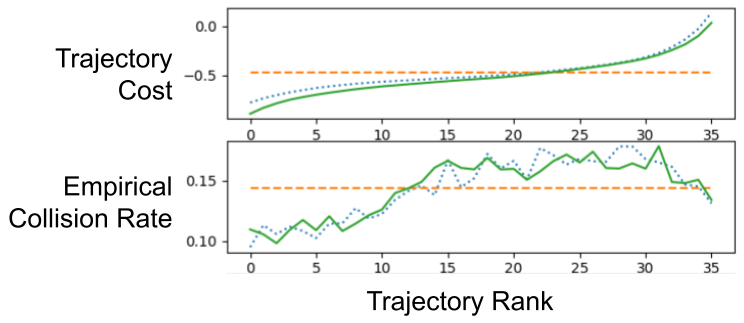}
\vspace*{-15pt}
\setlength{\belowcaptionskip}{-15pt}
\caption{\textbf{Trajectory  rank  versus  cost}.  The horizontal orange lines show the average cost of a randomly sampled trajectory.  The solid and dotted curves show the cost of a trajectory of a given rank, averaged across the \textbf{InD} dataset.}
\label{fig:rankingCosts}%
\end{figure} 

%% file: figure/collisionRateTable.tex
\begin{table}
\vspace{5mm}
\resizebox{\linewidth}{!}{%
\begin{tabular}{lcc|ll|l}
& Dataset & Com & \multicolumn{2}{c}{Top-$k$ Collision Rate (\%)} & \\
& & Agents & \multicolumn{2}{c}{[\% Change from $n=1$]} & $rand$ \\
& & & $k=1$ & $k=10$ & $traj$ \\
\hline
&                      & $n=1$ & 18.3 & 19.2 & 26.0\\ 
& \textbf{AutoCastSim} & $n=2$ & 16.2 [-11.2] & 15.8 [-17.3] & \\ 
& BlindCorner          & $n=4$ & 17.8 [-2.8] & 17.5 [-8.7] & \\
& RedLight             & $n=8$ & \textbf{7.9 [-57.0]} & 10.2 [-46.6] & \\
\hline
&                   & $n=1$ & 42.7 & 41.5 & 41.3\\ 
& \textbf{CoBEV-Sim} & $n=5$ & 36.7 [-14.1] & 36.9 [-11.3] & \\ 
& CrissCross        & $n=10$ & 35.5 [-16.9] & 36.8 [-11.5] & \\
&                   & $n=15$ & \textbf{30.0 [-29.7]} & 30.9 [-25.6] & \\
\end{tabular}
}
\setlength{\belowcaptionskip}{-15pt}
\caption{\textbf{Top-$k$ collision rate with $n$-agent communication}.  This table shows the collision rates of our method compared to a random sampling baseline.  The numbers in brackets show the percent change in collision rate as compared to the single-agent ($n=1$) performance.}
\label{tab:collisionRateTable}
\end{table}

%% file: conclusion.tex
\section{Conclusion}
This paper presents a novel method for addressing collision-free navigation for connected self-driving vehicles.  Namely, it uses a trajectory exchange mechanism to perform distributed validation of trajectory candidates for any number of neighboring vehicles.  The method exhibits several desirable traits, including being (1) distributed, (2) uncertainty-aware, and (3) bandwidth-efficient.  The method scales with the number of agents, reducing the collision rate up to $57\%$ with eight agents compared to the single-agent version. 

\textbf{Future Work}. Future work will further investigate the multi-agent setting and architecture.  By demonstrating favorable results for an extensible and interpretable neural architecture, this work laid a strong foundation for future multi-agent collaborative planning experiments.  Such experiments will explore the impact of localized sensor noise and degradation, reduced individual sensor capability, positioning noise, and variable network connectivity.  Each of these parameters adds realism to the multi-agent planning challenge, and future work will address their impact on the overall planning performance of the connected vehicle network.

\section{Acknowledgement}
\label{sec:acknowledgement}
\noindent This work was supported by ONR grant N00014-18-1-2829.